\pdfoutput=1

\documentclass{article}

\PassOptionsToPackage{numbers, compress}{natbib}

\usepackage[preprint]{neurips_2021}

\usepackage[utf8]{inputenc} 
\usepackage[T1]{fontenc}    
\usepackage{hyperref}       
\usepackage{url}            
\usepackage{booktabs}       
\usepackage{amsfonts}       
\usepackage{nicefrac}       
\usepackage{microtype}      
\usepackage[table,xcdraw]{xcolor}         
\usepackage{wrapfig}
\usepackage{amsthm,amsmath,amssymb}
\usepackage{mathrsfs}
\usepackage{graphicx}
\usepackage{subcaption}
\usepackage{multirow}
\usepackage{siunitx}
\usepackage{makecell}

\usepackage{kantlipsum}

\newcommand{\mrc}[2]{\multirowcell{#1}{#2}}

\title{BoolNet: Minimizing the Energy Consumption\\of Binary Neural Networks} 

\author{
	Nianhui Guo\thanks{Equal contribution.} \ \textsuperscript{1},
	Joseph Bethge\footnotemark[1] \ \textsuperscript{1}, 
	Haojin Yang\textsuperscript{1},
	Kai Zhong\textsuperscript{2}, \\
	\textbf{Xuefei Ning\textsuperscript{2},
		Christoph Meinel\textsuperscript{1}
		and Yu Wang\textsuperscript{2}} \\
	\textsuperscript{1} \ Hasso Plattner Institute, Germany\\
	\textsuperscript{2} \ Department of Electronic Engineering, Tsinghua University, China\\
	\texttt{\small \{nianhui.guo,joseph.bethge,haojin.yang,christoph.meinel\}@hpi.de}\\
	\texttt{\small \{zhongk19,nxf16\}@mails.tsinghua.edu.cn, yu-wang@tsinghua.edu.cn}
}

\begin{document}

\maketitle

\begin{abstract}
\label{sec:abstract}
Recent works on Binary Neural Networks (BNNs) have made promising progress in narrowing the accuracy gap of BNNs to their 32-bit counterparts.
However, the accuracy gains are often based on specialized model designs using additional 32-bit components.
Furthermore, almost all previous BNNs use 32-bit for feature maps and the shortcuts enclosing the corresponding binary convolution blocks, which helps to effectively maintain the accuracy, but is not friendly to hardware accelerators with limited memory, energy, and computing resources. 
Thus, we raise the following question:
\textit{``How can accuracy and energy consumption be balanced in a BNN network design?''}
We extensively study this fundamental problem in this work 
and propose a novel BNN architecture without most commonly used 32-bit components: \textit{BoolNet}.
Experimental results on ImageNet demonstrate that BoolNet can achieve $4.6\times$ energy reduction coupled with 1.2\% higher accuracy than the commonly used BNN architecture Bi-RealNet \cite{liu2018bi}.
Code and trained models are available at: \href{https://github.com/hpi-xnor/BoolNet}{https://github.com/hpi-xnor/BoolNet}.
\end{abstract}


\section{Introduction}
\label{sec:introcution}

The recent success of \emph{Deep Neural Networks} (DNNs) is like the jewel in the crown of modern AI waves. 
However, the large size and the high number of operations cause the current DNNs to heavily rely on high-performance computing hardware, such as GPU and TPU. 
Training sophisticated DNN models also results in excessive energy consumption and CO$_{2}$ emission, e.g., training the OpenAI's GPT-3 \cite{NEURIPS2020_1457c0d6} causes as much CO$_{2}$ emissions as 43 cars during their lifetime \cite{patterson2021carbon}.
Moreover, their computational expensiveness strongly limits their applicability on resource-constrained devices such as mobile phones, IoT devices, and embedded devices.
Various works aim to solve this challenge by reducing memory footprints and accelerating inference.
We can roughly categorize these works into the following directions:
network pruning \cite{han2015deep,han2015learning},
knowledge distillation \cite{crowley2018moonshine,polino2018model},
compact networks \cite{howard2017mobilenets,howard2019searching,sandler2018mobilenetv2,ma2018shufflenet,tan2019mnasnet}, and
low-bit quantization \cite{courbariaux2015binaryconnect,rastegari2016xnor,zhou2016dorefa,hubara2016binarized}.
From the latter, there is an extreme case, Binary Neural Networks (BNNs) (first introduced by \cite{courbariaux2016binarized}) that uses only 1 bit for weight and activation. 

\begin{figure}[]
\captionsetup[subfigure]{justification=centering}
\begin{center}
\begin{subfigure}[t]{0.3\linewidth}
    \vskip 0pt
   \centering
   \includegraphics[width=\linewidth]{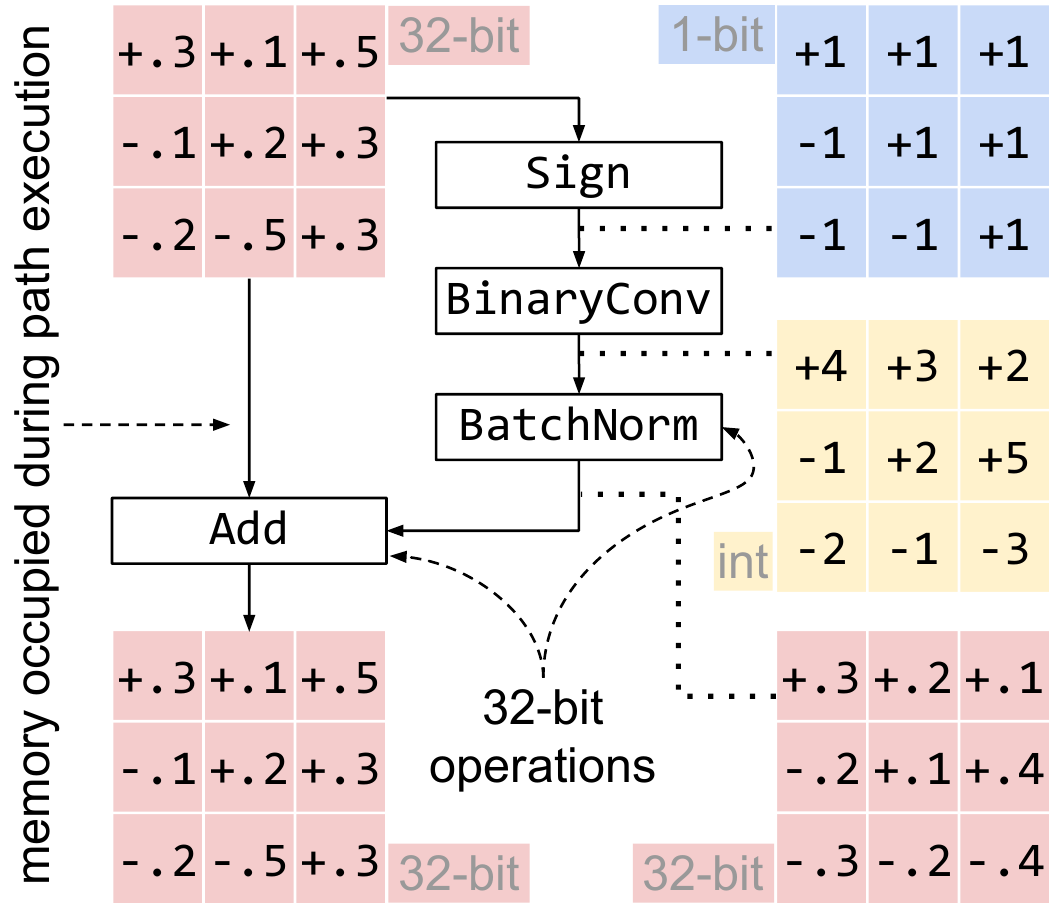}
   \caption{Design in previous work.}
   \label{fig:overview-previous}
\end{subfigure}
\hfill
\begin{subfigure}[t]{0.3\linewidth}
    \vskip 0pt
   \centering
   \includegraphics[width=\linewidth]{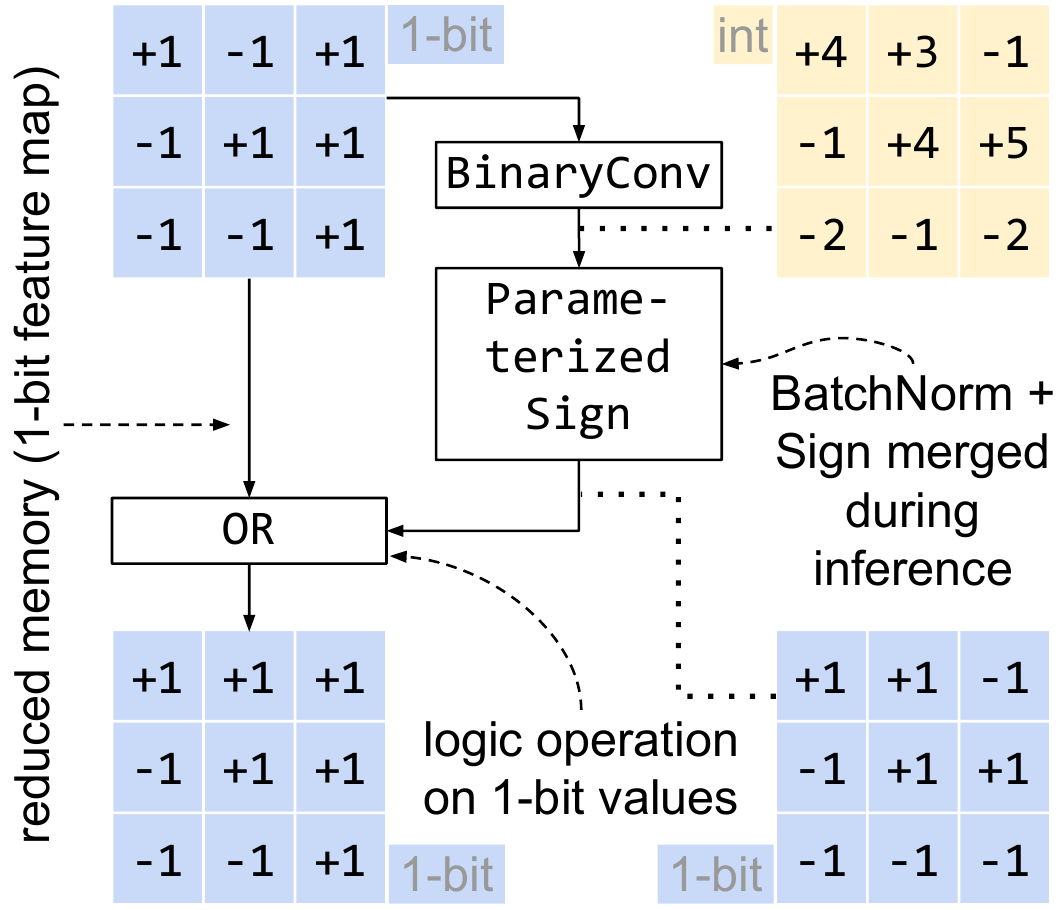}
   \caption{BoolNet design.}
   \label{fig:overview-boolnet}
\end{subfigure}
\hfill
\begin{subfigure}[t]{0.38\linewidth}
    \vskip 0pt
   \centering
   \setlength\tabcolsep{1.5pt}
\small
\begin{tabular}{ccccc}
\Xhline{4\arrayrulewidth}
\makecell{Method} & \makecell{\scriptsize Bitwidth\\\scriptsize(W/A/F)} &
\makecell{Energy\\(mJ)} & \makecell{Top-1\\Acc.} & \makecell{OPs\\($\cdot10^8$)} \\ \hline
\makecell{Bi-Real-\\Net \cite{liu2018bi}}   & 1/1/32                  & 3.90 & 56.4\% & 1.63 \\ \hline
\makecell{BoolNet\\(ours)}                  & 1/1/4                   & 0.84 & 57.6\% & 1.64 \\ \hline
\makecell{BaseNet\\(ours)}                  & 1/1/1                   & 0.61 & 48.9\% & 1.51 \\ \hline
\Xhline{4\arrayrulewidth}
\end{tabular}


   \caption{BoolNet reduces energy consumption by  ${4.6\times}$ compared to Bi-RealNet.}
   \label{fig:overview-reslts}
\end{subfigure}
\end{center}
\caption{The main differences between previous work and BoolNet.
BoolNet uses 1-bit feature maps and logic operations reducing memory requirements and the need for 32-bit operations.}
\label{fig:overview}
\end{figure}

As shown in the literature \cite{rastegari2016xnor}, BNNs can achieve 32$\times$ memory compression and up to 58$\times$ speedup on CPU, since the conventional arithmetic operations can be replaced by bit-wise \texttt{xnor} and \texttt{bitcount} operations.
However, BNNs suffer from accuracy degradation compared to their 32-bit counterparts.
For instance, XNOR-Net leaves an 18\% accuracy gap to ResNet-18 on ImageNet classification \cite{rastegari2016xnor}.
Therefore, recent efforts (analyzed in more detail in Section \ref{sec:related_work}) mainly focus on narrowing the accuracy gap, including
specific architecture design \cite{liu2018bi,bethge2019binarydensenet,bethge2020meliusnet,liu2020reactnet},
real-valued weight and activation approximation \cite{NIPS2017_b1a59b31,zhuang2019structured},
specific training recipes \cite{real2binICLR20},
a dedicated optimizer \cite{NEURIPS2019_9ca8c9b0},
leveraging neural architecture search \cite{bulat2020bats,zhao2020bars}
and dynamic networks \cite{bulat2020high}.
In the existing work, efficiency analysis usually only considers the theoretical instruction counts.
However, memory usage, inference efficiency and energy consumption, which are essential to practical applications, have received little attention.
Furthermore, \cite{fromm2020riptide} points out that the theoretical complexity is often inconsistent with the actual performance in practice and measurable performance gains on existing BNN models are hard to achieve as the 32-bit components in BNNs (such as BatchNorm, scaling, and 32-bit branches) become bottlenecks.
Using 32-bit information flow (e.g., 32-bit identity connections, 32-bit downsampling layers are equipped by almost all latest BNNs, see Figure \ref{fig:overview-previous}), and multiplication/division operations (in BatchNorm, scaling, average pooling etc.) significantly increase the memory usage and power consumption of BNNs and are thus unfriendly to hardware accelerators.
For these reasons, even if BNNs have achieved MobileNet-level accuracy with a similar theoretical number of OPs \cite{bethge2020meliusnet,real2binICLR20}, they still cannot be used as conveniently as compact networks \cite{howard2017mobilenets,howard2019searching,sandler2018mobilenetv2}.
%
%
%
 
In this paper, we extensively study the trade-off between BNN's accuracy and hardware efficiency.
We propose a novel BNN architecture: BoolNet, which replaces most commonly used 32-bit components (see Section \ref{sec:methodology}).
First, BoolNet uses binary feature maps in the network - by shifting the starting point of the shortcuts from the BatchNorm (BN) layer to the Sign function (see Figure \ref{fig:overview-boolnet}) - and uses Boolean functions instead of 32-bit additions to accumulate features.
Second, during inference, we fuse the BN layer into the Sign function through a lossless transformation, thereby effectively removing the MAdds brought by BN.
Other changes include removing components that require additional 32-bit multiplication/division operations: (1) PReLU, (2) average pooling, and (3) binary downsampling convolutions.
We further propose a \emph{Multi-slice strategy} to help alleviate the loss of representational capacity incurred by binarizing the feature maps and shortcut connections.
We show the effectiveness of our proposed methods and the increased energy efficiency of BoolNet with experiments on the ImageNet dataset \cite{deng2009imagenet}.
The results show the key benefit of BoolNet: a reasonable accuracy coupled with a higher energy efficiency over state-of-the-art BNNs (see Figure \ref{fig:overview-reslts} for a brief summary and Section \ref{sec:experiment} for more details).
The energy data is obtained through hardware accelerator simulation (see Section \ref{sec:energy_analysis} for details).
We summarize our main contributions as follows:
\begin{itemize}
  \item The first work studying the effects of 32-bit layers often used in previous works on BNNs.
  \item A novel BNN architecture BoolNet with minimal 32-bit components for higher efficiency.
  \item A Multi-slice strategy to alleviate the accuracy loss incurred by using 1-bit feature maps.
  \item State-of-the-art performance on the trade-off between accuracy and energy consumption with a $4.6\times$ lower power consumption than Bi-RealNet \cite{liu2018bi} and 1.2\% higher accuracy.
\end{itemize}


\section{Related Work}
\label{sec:related_work}

In recent years, \emph{Efficient Deep Learning} has become a research field that has attracted much attention.
Technical directions, such as,
compact network design \cite{howard2017mobilenets,howard2019searching,sandler2018mobilenetv2,zhang2018shufflenet,ma2018shufflenet},
knowledge distillation \cite{crowley2018moonshine,polino2018model},
network pruning \cite{han2015deep,han2015learning,li2016pruning,he2017channel}, and
low-bit quantization \cite{courbariaux2015binaryconnect,rastegari2016xnor,liu2018bi,liu2020reactnet,bethge2020meliusnet} are proposed for model compression and acceleration.
The efficient models have evolved from the earliest handcrafted designs to the current use of neural architecture search to search for the best basic block and overall network structure \cite{tan2019mnasnet,howard2019searching,tan2019efficientnet,radosavovic2020designing}.
The criterion of efficiency evaluation has also changed from instruction and parameter counts to more precise measurements of actual memory and operating efficiency on the target hardware \cite{cai2019once,cai2018proxylessnas}.

Binary Neural Networks were first introduced by Courbariaux et al. \cite{courbariaux2016binarized} and their initial attempt only evaluated on small datasets such as MNIST \cite{lecun-mnisthandwrittendigit-2010}, CIFAR10 \cite{krizhevsky2009learning} and SVHN \cite{netzer2011reading}.
The follow-up XNOR-Net \cite{rastegari2016xnor} proposes channel-wise scaling factors for approximating the real-valued parameters, which achieves 51.2\% top-1 accuracy on ImageNet.
However, there is an 18\% gap compared with its 32-bit counterpart, ResNet-18.
Therefore, recent efforts mainly focused on narrowing the accuracy gap.
WRPN \cite{mishra2018wrpn} shows that expanding the channel width of binary convolutions can obtain a better performance.
In ABC-Net \cite{NIPS2017_b1a59b31} and GroupNet \cite{zhuang2019structured}, instead of using a single binary convolution, they use a set of k binary convolutions (referred to as binary bases) to approximate a 32-bit convolution. 
This sort of method achieves higher accuracy but increases the required memory and number of operations of each convolution by the factor k.
Bi-RealNet \cite{liu2018bi} proposes using real-valued (32-bit) shortcuts to maintain a 32-bit information flow, which effectively improves the accuracy.
This design strategy became a standard for later work e.g., \cite{bethge2019binarydensenet,bethge2020meliusnet,liu2020reactnet}.
Martinez et al. \cite{real2binICLR20} propose using a real-valued attention mechanism and well-tuned training recipes to boost the accuracy further.
Thanks to the special architecture design, the recent MeliusNet \cite{bethge2020meliusnet} and ReActNet \cite{liu2020reactnet} achieve MobileNet-level accuracy with similar number of theoretical operations.
Other attempts, such as leveraging neural architecture search \cite{bulat2020bats,zhao2020bars} and dynamic networks \cite{bulat2020high}, show that those successful methods on regular real-valued networks are also effective for BNN.
Often, with improved accuracy, 32-bit components are used more frequently as well, such as PReLU and BatchNorm after each binary convolution \cite{liu2020reactnet}, real-valued attention module \cite{real2binICLR20} and scaling factors, etc.
On the contrary, efficiency analysis in the literature often only considers the theoretical operation number.
However, the memory usage and the actual energy consumption has received very little attention so far.


\section{BoolNet}
\label{sec:methodology}

In this section, we first revisit the latest BNNs and recap how they enhanced the accuracy by adding more 32-bit components (in Section \ref{subsec:standard binary neural networks with 32-bit components}). 
Afterwards, we propose to replace most commonly used 32-bit components from current BNN designs and instead use a fully binary information flow in the network (in Section \ref{subsec:replacing 32-bit componnets with boolean operations}).
However, abandoning 32-bit information flow results in a serious degradation of the representative capacity of the network.
Thus, we also present our strategies to restore the representative capacity (in Section \ref{subsec:boolnet with enhanced representation capacity}).
The focus on boolean operations and binary feature maps leads to the name of our network: \textbf{BoolNet}.
%

\subsection{Improving Accuracy with Additional 32-bit Components}
\label{subsec:standard binary neural networks with 32-bit components}
Recent works on BNNs have made promising progress in narrowing the gap to their 32-bit counterparts.
The key intention is to enhance the representative capacity by fully exploiting additional 32-bit components. 
However, such additional 32-bit components significantly reduce the hardware efficiency (as shown in \cite{fromm2020riptide} and further discussed in Section \ref{sec:energy_analysis}).
The following list summarizes the 32-bit components commonly used in the latest BNNs:
%
%
\begin{itemize}
    \item 
        The \textbf{channel-wise scaling factor} was first proposed by XNOR-Net \cite{rastegari2016xnor} for approximating the 32-bit parameters.
        It increases the value range of activation and weight.
    \item 
        Bi-RealNet \cite{liu2018bi} proposes to use a \textbf{32-bit shortcut} for enclosing each binary convolution. 
        The key advantage is that the network can maintain an almost completely 32-bit information flow (cf. Figure \ref{fig:block-baseline}).
    \item
        XNOR-Net \cite{rastegari2016xnor} uses \textbf{32-bit 1$\times$1 downsampling} convolutions, which is also used by most subsequent methods \cite{liu2018bi,real2binICLR20,bethge2020meliusnet}.
        \cite{bethge2019binarydensenet} shows that this simple strategy can achieve about 3.6\% Top-1 accuracy gains on ImageNet based on a binary ResNet-18 model.
        
    \item
        \cite{real2binICLR20,bulat2020bats,bulat2020high} show that \textbf{PReLU activation} effectively improves accuracy of BNNs.
        ReActNet \cite{liu2020reactnet} constructs the RPReLU activation function and uses it before every sign function.
    \item
        Real-to-Binary Net \cite{real2binICLR20} reuses the 32-bit activation after BN through squeeze and excitation (SE) \textbf{attention mechanism}.
        This module can adaptively re-scale the outputs of each binary convolution but needs additional 32-bit operations.
\end{itemize}

Although these techniques can effectively improve the accuracy, 
they increase the number of 32-bit values and floating point operations, making them not particularly efficient on hardware accelerators.
They are closer to mixed-precision neural networks rather than being highly efficient binary neural networks, as one might expect. 


\subsection{BaseNet: Replacing 32-bit Components with Boolean Operations}
\label{subsec:replacing 32-bit componnets with boolean operations}
\begin{figure}[]
\captionsetup[subfigure]{justification=centering}
\begin{center}
\begin{subfigure}[t]{0.41\linewidth}
   \centering
   \includegraphics[scale=0.6]{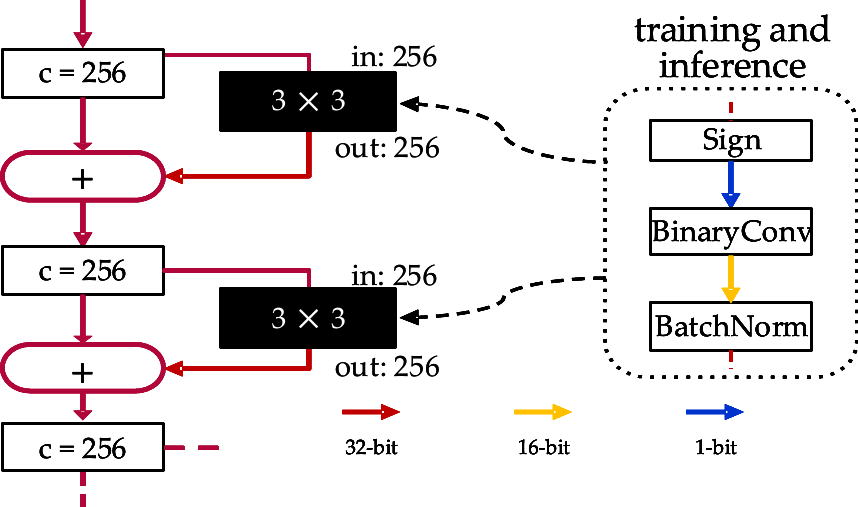}
   \caption{Typical binary basic block with 32-bit shortcuts and Batch Normalization layer.}
   \label{fig:block-baseline}
\end{subfigure}
\hfill
\begin{subfigure}[t]{0.57\linewidth}
   \centering
   \includegraphics[scale=0.6]{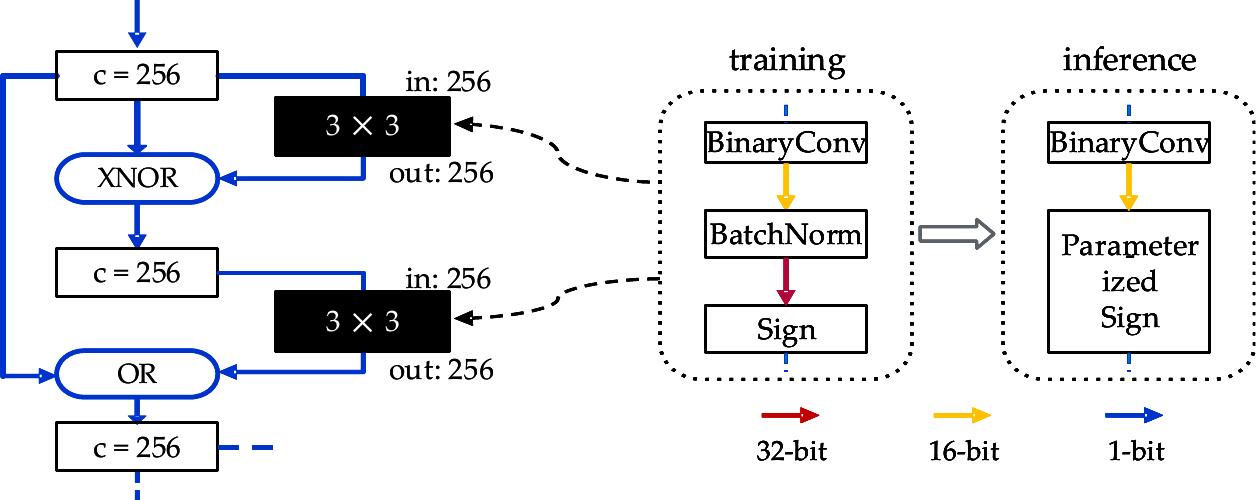}
   \caption{Our binary block design with logic shortcuts without 32-bit operations. $c$ indicates the number of channels.}
   \label{fig:block-boolnet}
\end{subfigure}
\end{center}
\caption{Comparison between a conventional binary convolution block with 32-bit shortcuts (a) and our proposed BoolNet convolution block with 1-bit logic shortcuts (b).}
\label{fig1: shortcut comparision}
\end{figure}

To better balance accuracy and efficiency, we rethink the additional 32-bit components (Batch Normalization, 32-bit feature maps, scaling factors and PReLU) elaborated in the previous section and propose to replace them with boolean operations.
We further propose a new basic convolution block without 32-bit operations, as shown in Figure \ref{fig:block-boolnet}, where we rearranged the order of convolution basic block as \{BinaryConv-BatchNorm-Sign\}, so that all feature maps are binary.
These general changes constitute our BoolNet baseline, in short \emph{BaseNet}.

%
%


\subsubsection{Integrating BatchNorm into Sign Function}
\label{subsec:integrating_BN}
Most studies on binary neural architecture design have kept the 32-bit BatchNorm (BN) layer in both the training and testing stages \cite{hubara2016binarized,rastegari2016xnor,liu2018bi,liu2020reactnet,bethge2020meliusnet}. 
However, using a 32-bit BN right after the 1-bit convolution layer decreases the computational efficiency on hardware, using more memory and energy.
Thus, in the following we propose to fuse the BN layer into the Sign function during the inference stage.

During the training phase, the batch normalization layer normalizes feature maps with an running mean $\mu$ and a running variance $\sigma$. 
For inference, it utilizes the constant statistic mean and variance instead, which in result can be reformulated as a linear process, expressed as:
\begin{equation}
    \begin{aligned}
        y_i=&\,\,\gamma\frac{x_i-\mu}{\sqrt{||\sigma ^2+\epsilon ||}}+\beta =\frac{\gamma}{\sqrt{||\sigma ^2+\epsilon ||}}x_i+\left( \beta -\frac{\gamma \mu}{\sqrt{||\sigma ^2+\epsilon ||}} \right)
    \label{equa:batch norm}
    \end{aligned}
\end{equation}
where $x_i$ and $y_i$ represent the N-dimensional input and output of a BN layer.  
$\gamma$ and $\beta$ are trainable scale and shift parameters, which are constant during the inference.
$||\dots||$ is the absolute function. 
We can therefore simplify the formula as follows:
\begin{equation}
    \begin{aligned}
        y_i=&\,\,ax_i+b\,\,=\,\,a\left( x_i+\frac{b}{a} \right) = a\left( x_i + c\right) \,\,,
    \label{equa:bn simplified}
    \end{aligned}
\end{equation}
where $a$, $b$, and $c$ denote constants in the formula.
By transforming $a$ into its sign and its absolute value, we have
\begin{equation}
    \begin{aligned}
        y_i=&\,\,||a||\circledast \text{Sign}\left( a \right) \odot \left( x_i+c \right),
    \label{equa:transformed BN}
    \end{aligned}
\end{equation}
As arranged in our basic block, Equation (\ref{equa:transformed BN}) is followed by a sign function, and $\text{Sign}(y_i)$ only depends on $\text{Sign}(a)$ and $(x_i+c)$.
We thus derive a parameterized sign function as:
\begin{equation}
    \begin{aligned}
        \text{Sign}(y_i) 
        =&\,\,\text{XNOR}(\text{Sign}(a),\,\text{Sign}(x_i+c))
    \label{equa:sign of BN}
    \end{aligned}
\end{equation}
We further replace $\odot$ by using XNOR operator so that only bitwise operations are adopted in the inference.

\subsubsection{1-bit Logic Shortcuts}
\label{subsec:logic shortcuts}
The residual shortcut is usually a 32-bit branch 
which branches off after the BatchNorm (BN) and pointwise addition in previous work \cite{liu2018bi,liu2020reactnet,bethge2020meliusnet}.
We modify the residual shortcut in two aspects:
(i) We shift the starting point of the shortcut connection from the output of BN to the output of the Sign function.
(ii) We utilize the logic operators XNOR and OR for merging the binary features to the consecutive block (instead of 32-bit addition).
Based on this novel shortcut design, called \textbf{Logic Shortcuts}, the feature maps in each stage of the network is completely binary without 32-bit operations.
It reduces the memory consumption of the intermediate feature maps by 32$\times$ and is the first binary residual structure proposed for BNNs to the best of our knowledge.

Although boolean operators can fulfill the needs of fusing binary information branches, they are not inherently differentiable.
To allow our network with boolean operators to be trained using back-propagation, we replace XNOR and OR in the training stage with the following differentiable terms:
\begin{equation}
    \text{XNOR}\left( x',\,y' \right) =x\cdot y~~~~~~~~~~~~~\text{OR}\left( x',\,y' \right) =2\cdot \textrm{Min}\left( 1, \frac{x+y}{2}+1 \right)-1
\end{equation}
where $x,\,y \in \{-1, +1\}$ denote the binary variables during training (and $x',\,y' \in \{0, 1\}$ during inference).
This allows us to convert them back to logic operators during inference loss-free.

In summary, our proposed basic block (see Figure \ref{fig:block-boolnet}) maximizes efficiency by using only 1-bit operations during inference and uses two different logic shortcuts based on XNOR and OR. 
This is contrary to conventional BNN blocks \cite{liu2018bi,liu2020reactnet}, which use 1-bit only for convolution layers, whereas other components are 32-bit or 16-bit (cf. Figure \ref{fig:block-baseline}), 

\subsubsection{Further Reducing 32-bit Operations}
\label{subsec:reducting fp ops}
We rarely use the PReLU activation function, which is commonly used in most literature \cite{liu2018bi,real2binICLR20} and brings a lot of extra overhead to the hardware implementation (it is only used once before the final dense layer).
We also decided not to use scaling factor as suggested by \cite{liu2018bi,bethge2019binarydensenet}.
Furthermore, we binarize the 1$\times$1 downsampling convolution, which is usually kept full-precision in previous methods \cite{liu2018bi,real2binICLR20} without the severe accuracy loss described in previous work \cite{rastegari2016xnor,liu2018bi}.
This further reduces the number of 32-bit operations and 32-bit parameters in BoolNet, but due to space limitations, we discuss the details on, alternatives to, and results of these changes in the supplementary material.
There are two components using 32-bit operations and parameters in previous work, which are kept in 32-bit in BoolNet: the first convolution and the last dense layer.
Directly replacing them with binary versions leads to a severe accuracy loss \cite{rastegari2016xnor}, thus we leave the investigation of alternatives for these special cases for future work.

\subsection{BoolNet: Enhancing Binary Information Flow}
\label{subsec:boolnet with enhanced representation capacity}

%
\begin{figure}[]
\captionsetup[subfigure]{justification=centering}
\begin{center}
\begin{subfigure}[t]{0.29\linewidth}
   \centering
   \includegraphics[scale=0.38]{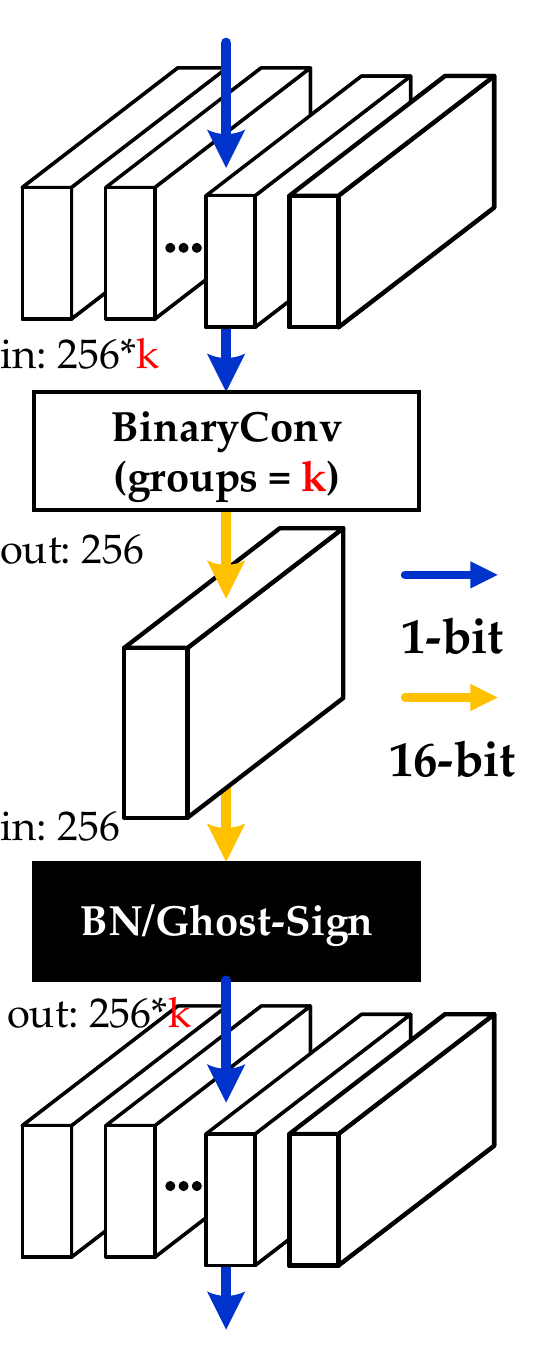}
   \caption{Multi-Slices binary convolution}
   \label{fig:boolnetv2-msbconv}
\end{subfigure}
\hfill
\begin{subfigure}[t]{0.29\linewidth}
   \centering
   \includegraphics[scale=0.38]{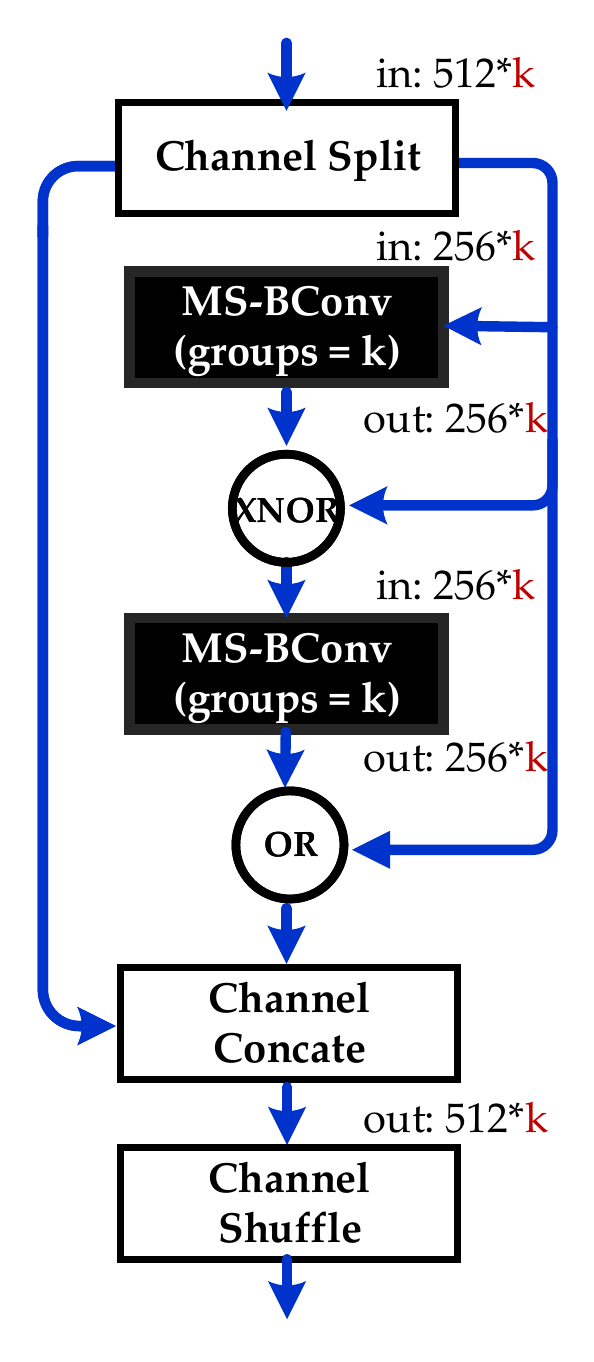}
   \caption{BoolNet basic block}
   \label{fig:boolnetv2-basicblock}
\end{subfigure}
\hfill
\begin{subfigure}[t]{0.39\linewidth}
   \centering
   \includegraphics[scale=0.38]{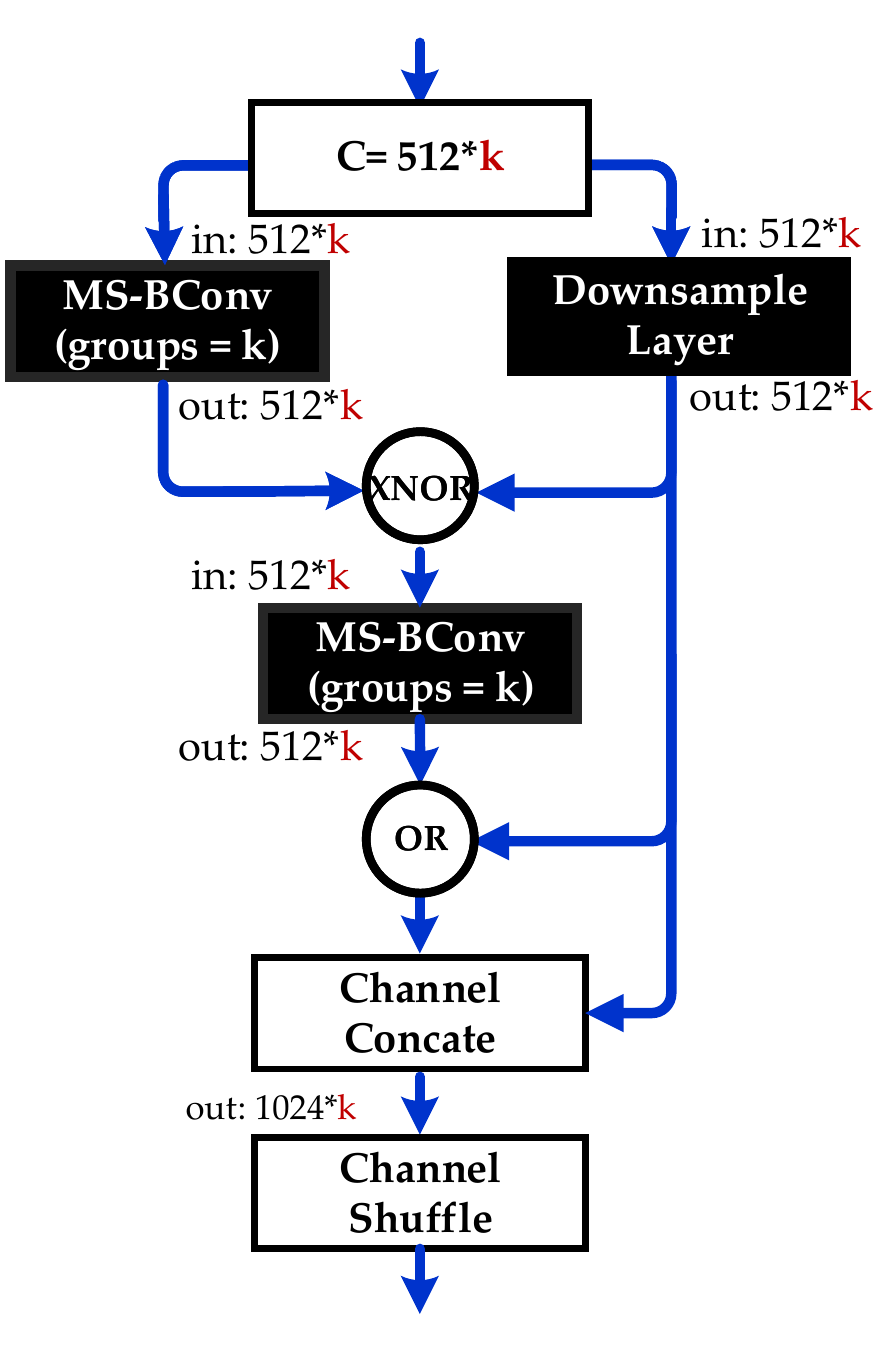}
   \caption{BoolNet downsample block}
   \label{fig:boolnetv2-downsample}
\end{subfigure}
\end{center}
\caption{Detail Architecture of BoolNet. To enhance the information flow, we modify the baseline architecture from two aspects: a) Reducing information loss through multi-slices binary convolution. b) Strengthening information propagation by features reusing.}
\label{fig:boolnetv2}
\end{figure}
%


The network design changes explained in the previous section, constitute our BoolNet baseline, called \emph{BaseNet}.
Although it uses a completely binary information flow which minimizes the energy and memory consumption, the representative capacity of BaseNet is drastically degraded compared to its 32-bit counterparts.
To counter this reduction of representative capacity, we propose the following two ideas, which constitute our proposed \textbf{BoolNet}.

\textbf{Multi-slices Binary Convolution}.
Instead of using a single 1-bit value for each 32-bit value in a regular BNN, our multi-slice strategy proposes of using a set of $k$ 1-bit values.
The key intention is to reduce the information loss caused by the sign function. 
We consider the typical binarization process $\text{Sign}(x_i, \text{zero-point})$ as a special case of single-slice numerical projection. 
Thus, we propose a multi-slice projection strategy for binary convolution to retain more relative magnitude information.
Specifically, we redesign sign function as follows:
\begin{equation}
    x_i^b = \text{Sign}(x_i, b_n),
    \label{equa:x_b_slice}
\end{equation}
where $b_n$ indicates a set of constant bias:
\begin{equation}
    b_n={\frac{\pm 2n}{k}}, \text{where}\,\,n = 0, 1, ...\,,k/2
    \label{equa:b_n}
\end{equation}
We adopt $b_n$ to conveniently expand the channel dimension to enhance the capacity of the binary feature map.
If $n=0,k=1$, Equation (\ref{equa:b_n}) degenerates to the ordinary sign function.
In Equation (\ref{equa:x_b_slice}), $x_i^b$ denotes the binary projection output with the dimension of $[N, C*k, H, W]$, which will be fed into the subsequent binary convolution layer. 
The constant $k$ also denotes the group number of the convolution.
That is, by setting the number of groups to $k$ in each convolution, the overall amount of parameters and operations of each convolution is unchanged.
Motivated by FReLU \cite{ma2020funnel}, we enhance the first multi-slices projecting module, after the input convolution of network, with a \textbf{Local Adaptive Shifting} module. 
This module consists of a depth-wise $3\times3$ convolution and a batch normalization layer and is able to adaptively change the zero points of each pixel, in a light-weight manner.
For simplicity, the multi-slices binary convolution is referred to as \textbf{MS-BConv}, subsequently.
Figure \ref{fig:boolnetv2-basicblock} shows the detailed block design of MS-BConv.

\textbf{Strengthening The Information Propagation in BoolNet}.
The layer-by-layer feature extraction and accumulation mechanism are key reasons deep neural networks have strong representative capacity.
Unlike typical residual shortcuts, which accumulates information from shallow to deep based on addition operation, 
logic shortcuts using boolean operators such as XNOR and OR can only represent True and False states, making them difficult to accumulate and propagate information.
To alleviate this bottleneck, we strengthen feature propagation by reusing features.
In ShuffleNet-V2 \cite{2018ShuffleNet}, the input tensor is divided into two equal parts, the first half is used for feature extraction, 
and the other half is directly copied and concatenated with the extracted features.
Inspired by its characteristics of information fusion and retention, we use a similar method to enhance the information retention capability of the BoolNet block.
As demonstrated in Figure \ref{fig:boolnetv2-basicblock}, the feature extraction branch consists of two MS-BConv modules with logic shortcuts, and the other branch remains identity. 
Two branches are concatenated and followed by channel shuffle, ensuring that the features from different layers are uniformly distributed.
Figure \ref{fig:boolnetv2-downsample} shows the downsampling block design of BoolNet, where no channel splitting is required, and it doubles the number of channels in the output.
Changing this information accumulation mechanism constitutes our proposed \textbf{BoolNet} over the \emph{BaseNet} (as referred to in Section \ref{sec:experiment}).

\subsection{Training with Progressive Weight Binarization}
\label{sec:progressive-weight-binarization}
Though we intend to build highly efficient BNNs with fully binary information flow, 
this strategy make the network more sensitive to weight initialization during training. 
Traditional methods have tried alleviate similar problem through two-stage training \cite{real2binICLR20,liu2020reactnet}, which makes training more complicated. 
In this paper, we adopt a progressive binarization technique based on the traditional Hardtanh-STE method \cite{courbariaux2016binarized}. This can be viewed as a smooth version of previous multi-stage training. 
Specifically, in the training phase, a differentiable function $F(x)$ is used to replace sign function.
During the forward, the slope of this function is adjusted by a single scalar $\lambda$. 
As the slope shrinks, the weight gradually changes from 32-bit to 1-bit.
During backward propagation, we approximate $F(x/\lambda)$ with ${F(x/1)}$, which escapes BoolNet from the gradient vanishing as $\lambda$ decreases. 
In the testing phase, we use traditional sign function for inference.
The whole process can be formulated as:
\begin{equation}
F(x, \lambda) = {\mathop {\lim}\limits_{\lambda \to 0}} \ \text{Hardtanh}\left(\frac{x}{\lambda}\right) \simeq \text{Sign}(x).
\label{equa:progressive binarizing}
\end{equation}
To smooth the weight binarization process, we schedule ${\lambda}$ during training with an exponential decay strategy ${\lambda}_t = {\sigma}^{(t)}$, where ${\sigma}<1$ is the exponential decay rate of $\lambda$.



\section{Experiments}
\label{sec:experiment}

We use the task of image classification on the ImageNet \cite{deng2009imagenet} dataset as our main means of evaluation.
In the following section, we first present the training details for our experiments. 
Afterwards, we study the effects of our proposed network design changes, (in Section \ref{sec:ablation-networks}) and the \emph{Multi-slice} convolution (in Section \ref{sec:ablation-slices}) and analyze the energy consumption of BoolNet and other recent work on BNNs (in Section \ref{sec:energy_analysis}) and compare our model accuracy to state-of-the-art BNN models (in Section \ref{sec:sota_comparison}).

\subsection{Training Details}
\label{sec:training_details}

Our general training strategy and hyperparameters are mostly based on \cite{bethge2020meliusnet}, the exact hyperparameters, training details and training code are available in the supplementary material.


As an alternative to the two-stage training approach, as described in \cite{liu2020reactnet,real2binICLR20}, we proposed progressive weight binarization (see Section \ref{sec:progressive-weight-binarization}, Equation \ref{equa:progressive binarizing}).
In the following experiments, we used $\sigma=0.965$ and thus $\lambda=0.965^t$, with $t$ being the number of iterations divided by 1000 (i.e. $\lambda$ is multiplied by $0.965$ every $256000$ samples).
Note, that the progressive weight binarization is replaced by a regular sign function during the validation pass.
The two stage training strategy aims to provide a good initialization for a BNN training, by first training a model with 1-bit activations/32-bit weights and weight decay of $10^{-5}$, and use it to initialize the training of a 1-bit activations/1-bit weights model.
We tested the effect of both strategies with a plain ResNet-like model with binary feature maps and our proposed Logic Shortcuts on ImageNet.
The two-stage training (trained 60 epochs in each stage - a total of 120 epochs) achieved 49.60\% accuracy.
Our progressive weight binarization achieves 48.39\% when training for \textbf{60} epochs, but achieves 50.19\% when training for \textbf{120} epochs.
Thus we deduce that our training strategy effectively removes the need for a two stage training (based on a similar total training time) and leads to a similar or better result.

\subsection{Ablation on Network Design}
\label{sec:ablation-networks}

\setlength\tabcolsep{2pt}
\begin{table}[]
\caption{Our ablation study on ImageNet \cite{deng2009imagenet} regarding accuracy, number of 32-bit operations (FLOPs), 1-bit operations (BOPs), and model size.
We \textbf{highlighted} the positive effects of \emph{Logic Shortcuts}, \emph{Local Adaptive Shifting}, and \emph{Multi-slice Convolution} ($k$ denotes the number of slices).}
\begin{center}
\scriptsize
\rule{14cm}{1pt}
\begin{tabular}{l|cccccc|cccccc}
                          & \multicolumn{6}{c|}{BaseNet} & \multicolumn{6}{c}{BoolNet} \\
 \multicolumn{1}{c|}{Network Configuration (k=1)} &
  \makecell{Top 1\\ Acc.}        & \makecell{Top 5\\ Acc.}       & \makecell{FLOPs\\($\cdot10^8$)} &
  \makecell{BOPs\\($\cdot10^9$)} & \makecell{OPs\\($\cdot10^8$)} & \makecell{Model\\ Size} &
  \makecell{Top 1\\ Acc.}        & \makecell{Top 5\\ Acc.}       & \makecell{FLOPs\\($\cdot10^8$)} &
  \makecell{BOPs\\($\cdot10^9$)} & \makecell{OPs\\($\cdot10^8$)} & \makecell{Model\\ Size} \\ \hline
Baseline (no shortcuts)       & 46.26\% & 70.84\% & 1.23 & 1.76 & 1.51 & 3.49 MB & -       & -       & -        & -        & -        & -       \\
+ Logic Shortcuts (XNOR/OR)   & \textbf{48.60\%} & \textbf{72.79\%} & 1.23 & 1.76 & 1.51 & 3.49 MB & 49.92\% & 74.17\% & 1.23 & 2.01 & 1.55 & 3.71 MB \\
+ Local Adaptive Shifting     & 48.83\% & 73.19\% & 1.26 & 1.76 & 1.53 & 3.49 MB & \textbf{51.51\%} & \textbf{75.41\%} & 1.26 & 2.01 & 1.57 & 3.71 MB
\end{tabular}
\rule{14cm}{1pt}
\small
\setlength\tabcolsep{2.8pt}
\begin{tabular}{c|cccccc|cccccc}
\multicolumn{1}{l|}{} & \multicolumn{6}{c|}{BaseNet (with Logic Shortcuts)}           & \multicolumn{6}{c}{\scriptsize BoolNet (with Logic Shortcut and Local Adaptive Binarization)}                                  \\
k &
  \makecell{Top 1\\ Acc.}        & \makecell{Top 5\\ Acc.}       & \makecell{FLOPs\\($\cdot10^8$)} &
  \makecell{BOPs\\($\cdot10^9$)} & \makecell{OPs\\($\cdot10^8$)} & \makecell{Model\\ Size} &
  \makecell{Top 1\\ Acc.}        & \makecell{Top 5\\ Acc.}       & \makecell{FLOPs\\($\cdot10^8$)} &
  \makecell{BOPs\\($\cdot10^9$)} & \makecell{OPs\\($\cdot10^8$)} & \makecell{Model\\ Size} \\ \hline
1                     & 48.60\% & 72.79\% & 1.23 & 1.76 & 1.51 & 3.49 MB & 51.51\% & 75.41\% & 1.26 & 2.01 & 1.57 & 3.71 MB \\
4                     & \textbf{52.15\%} & \textbf{75.89\%} & 1.23 & 2.01 & 1.55 & 3.56 MB & \textbf{54.45\%} & \textbf{77.83\%} & 1.26 & 2.50 & 1.65 & 3.84 MB \\
8                     & 52.51\% & 76.34\% & 1.23 & 2.35 & 1.60 & 3.65 MB & -       & -       & -        & -        & -        & -      
\end{tabular}
\rule{14cm}{1pt}
\label{tab:ablation-combined}
\end{center}
\end{table}

For brevity, we refer to our BoolNet baseline (consisting of our changes described in Section \ref{subsec:replacing 32-bit componnets with boolean operations}) as \emph{BaseNet}.
When we apply our changes regarding the information propagation described in Section \ref{subsec:boolnet with enhanced representation capacity}, we refer to it as \emph{BoolNet}.
In the following section, we study the effects of all of our proposed network design changes, in particular of the \emph{Logic Shortcut} (see Section \ref{subsec:logic shortcuts}) and the \emph{Local Adaptive Shifting} module (see Section \ref{subsec:boolnet with enhanced representation capacity}) on the ImageNet dataset.
Our results (see upper half of Table \ref{tab:ablation-combined}) show, that adding \emph{Logic Shortcuts} to a plain BaseNet (without shortcuts) to accumulate 1-bit features with XNOR and OR increases accuracy by 2.4\% with minimal extra cost.
We infer that such shortcuts can be a suitable replacement for the addition that were used to accumulate 32-bit features in previous BNNs and use them in all our network designs.
However, the \emph{Local Adaptive Shifting} module is only effective for our proposed BoolNet (providing an accuracy increase of 1.59\%) and does not provide a benefit for a BaseNet-style network (accuracy is increased by only 0.23\%) compared to the extra cost.



\subsection{Ablation on the Multi-slice Convolution}
\label{sec:ablation-slices}

We also evaluated whether using Multi-slice Convolutions (see Section \ref{subsec:boolnet with enhanced representation capacity}) can reduce the accuracy loss caused by using 1-bit feature maps ($k$ denotes the number of slices).
Our results on ImageNet (see lower half of Table \ref{tab:ablation-combined}) show, that using $k=4$ increases accuracy significantly for our BaseNet (3.55\%) and BoolNet (2.94\%) architectures.
Although the convolutions used throughout the network use a number of groups equal to $k$ to keep the required parameters and operations constant, operations and parameters are still slightly increased in the $1\times1$ convolution in the downsampling branch which uses all channels.
Overall $k=4$ leads to a slight increase of operations (compared to $k=1$), however is still significantly lower than compared to previous work \cite{real2binICLR20, liu2018bi}.
However, further increasing $k$ to $k=8$ only slightly improves accuracy (by 0.36\%), but again increases operations and parameters in the downsampling branch.
Therefore, $k=4$ provides the best trade-off, which is further proven in the following section.
Furthermore, using the Multi-slice strategy allows us to use a downsampling branch \emph{without 32-bit components without accuracy degradation} (using 1-bit 1$\times$1 convolutions) and use this design in our comparison to state-of-the-art.
(Due to space limitations, the details on our downsampling branch design are in the supplementary material.)


\subsection{Energy Consumption Evaluation}
\label{sec:energy_analysis}
This section evaluates the energy consumption of BoolNet and several classic BNN architectures through hardware simulation.
We design five accelerators for five BNNs in RTL language, and the power and area of computing circuits are given by Design Compiler (DC) simulation with TSMC 65nm process and 1GHz clock frequency.
We further evaluate the energy consumption of on-chip SRAM access and off-chip DRAM access by using CACTI 6.5 \cite{hp_labs:cacti}, and the power calculator of DDR provided by Micron \cite{Micron}.
The above components sum the overall energy consumed by a single inference pass.

\begin{figure}[]
\captionsetup[subfigure]{justification=centering}
\begin{center}
\begin{subfigure}[t]{0.59\linewidth}
    \vskip 0pt
   \centering
   \setlength\tabcolsep{3pt}
\small
\begin{tabular}{c|c|c|c|c|c}
        Operation       & \thead{Power\\($mw$)}        & \thead{Area\\($um^2$)}     &    Operation         & \thead{Power\\($mw$)} & \thead{Area\\($um^2$)} \\ \hline
BConv            & $108.8$            & $131737$                      & Int8 Conv(1/8) & $504$       & $836269$                      \\ \hline
1-bit Agg      & $1.4$              & $2150$                        & Int Agg     & $43.5$      & $53238$                       \\ \hline
16-bit Sign           & $1.4$              & $7956$                        & 32-bit Sign         & $3.3$       & $13548$                       \\ \hline
\thead{32-bit\\ RPReLU} & $137.6$              & $310671$                          & Int8 BN            & $50.1$      & $274606$                      \\ \hline
\end{tabular}
   \caption{Energy consumption and circuit area of different operations (see Section \ref{sec:energy_analysis} for a detailed explanation).}
   \label{tab:impact_reduced_real_valued_components}
\end{subfigure}
\hfill
\begin{subfigure}[t]{0.4\linewidth}
    \vskip 0pt
   \centering
   \includegraphics[width=\linewidth]{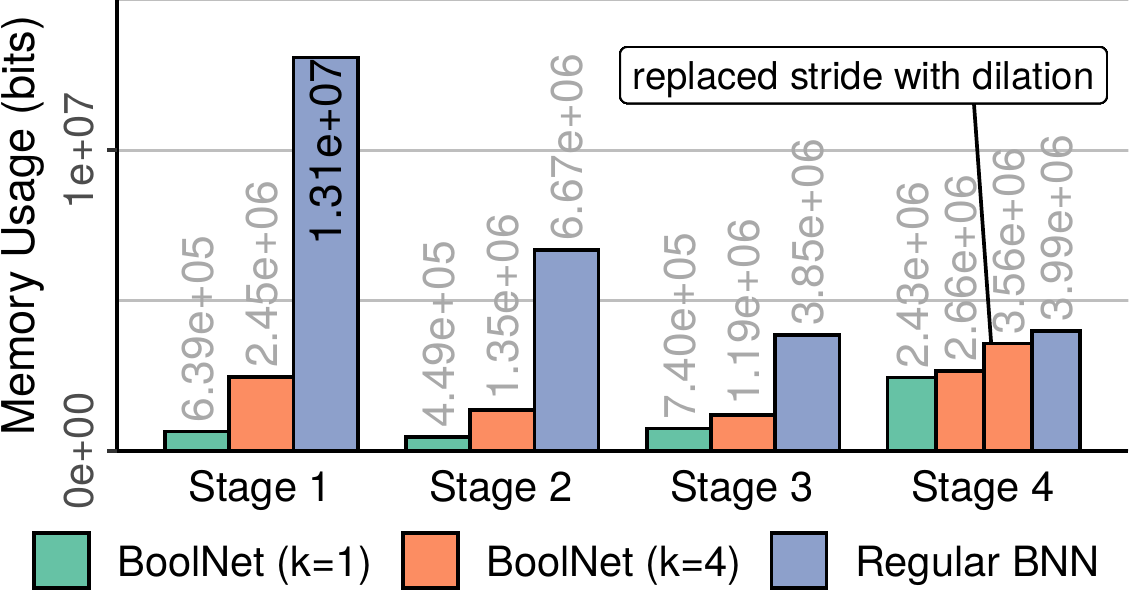}
   \caption{Memory usage comparison between blocks of different stages.}
   \label{fig:memory-graph}
\end{subfigure}
\end{center}
\caption{A theoretical memory usage comparison of one convolution block between BoolNet and previous work.
Actual numbers can differ during implementation, but BoolNet shows significantly lower memory usage, especially in early stages, even when using our Multi-slice strategy with $k=4$.}
\label{fig:memory}
\end{figure}

Memory access and computation are the primary factors that affect energy consumption of a hardware accelerator.
However, in the existing BNNs, efficiency analysis only considers the theoretical instruction counts \cite{liu2018bi,real2binICLR20,bethge2020meliusnet,liu2020reactnet} while the impact of memory access has been neglected.
A theoretical analysis (see Figure \ref{fig:memory-graph}) of the required memory shows that the total memory by BoolNet is much lower than previous BNNs, especially during the earlier stages of the network.
This analysis also shows that using dilation in the last stage of BoolNet still uses less memory for convolution blocks than in previous BNNs.
Our energy evaluation results (see Figure \ref{fig:sota-energy-graph}) show that the energy consumption of computing units accounts for a small proportion in the whole calculation. 
Our design achieves higher energy efficiency due to a lower memory access. 
In other BNNs, preserving and reading 32-bit feature maps drastically increase energy consumption.
Since the overall memory usage of BoolNet is minimal, it requires much less DRAM access than the others. 
Generally, DRAM has much higher power consumption than SRAM.

Furthermore, the energy consumption of some commonly used components is shown in Figure \ref{tab:impact_reduced_real_valued_components}.
For instance, the energy consumption of Int8 downsampling convolution is 37$\times$ larger than binary downsampling\footnote{37=504$\times$8/108.8, where Int8 Conv has only 1/8 of the parallel capability of BConv.}.
The \emph{Logic Shortcut} aggregation is 31$\times$ more energy efficient than additive aggregation.
Surprisingly, 32-bit PReLU consumes 26\% more energy than a binary convolution, Int8 BN consumes about half of a binary convolution, and those two components are commonly used in conjunction with binary convolutions in previous BNNs.
More implementation and evaluation details can be found in supplementary materials.

\subsection{Comparison to State-of-the-Art BNNs}
\label{sec:sota_comparison}

For our comparison to state-of-the-art BNNs, we replaced the Cross-Entropy loss with a knowledge distillation approach, based on the implementation of \cite{liu2020reactnet} with a 32-bit ResNet-34 \cite{he2016deep} as the teacher model and train the models for 80 or 90 epochs instead of 60 epochs.
(Due to limited hardware resources, we were not able to choose a longer training time, but suspect increasing the training time, e.g. to 120 epochs, could improve the results.)

\begin{figure}[]
\captionsetup[subfigure]{justification=centering}
\begin{center}
\vfill
\begin{subfigure}[t]{0.57\linewidth}
   \vskip 0pt
   \centering

\small
\begin{tabular}{ccccc}
\Xhline{4\arrayrulewidth}
Methods   & \makecell{Bitwidth\\\scriptsize(W/A/F)} & \makecell{\scriptsize Energy\\\scriptsize Consumption} & \makecell{Top-1 \\Acc.}  & \makecell{OPs\\($\cdot10^8$)}  \\ \hline
ReActNet (Bi-Real) \cite{liu2020reactnet}    & 1/1/32   & 3.93mJ            & 65.9\%            & 1.63            \\
Bi-RealNet \cite{liu2018bi}                  & 1/1/32   & 3.90mJ            & 56.4\%            & 1.63            \\  
XNOR-Net \cite{rastegari2016xnor}            & 1/1/32   & 1.92mJ            & 51.2\%            & 1.59            \\ \hline  
BoolNet$^*$, k=4 (ours)                           & 1/1/4    & 1.18mJ            & 59.6\%            & 1.76            \\  
BoolNet, k=4 (ours)                               & 1/1/4    & 0.84mJ            & 57.6\%            & 1.64            \\  
BaseNet, k=4 (ours)                               & 1/1/4    & 0.74mJ            & 55.1\%            & 1.54            \\
BaseNet, k=1 (ours)                               & 1/1/1    & \textbf{0.61mJ}   & 48.9\%            & 1.51            \\
\Xhline{4\arrayrulewidth}
\end{tabular}
   \caption{The advantage of BoolNet is reduced energy consumption.
   }
   \label{tab:sota-table}
\end{subfigure}
\hfill
\begin{subfigure}[t]{0.42\linewidth}
   \vskip 0pt
   \centering
   \includegraphics[width=\linewidth]{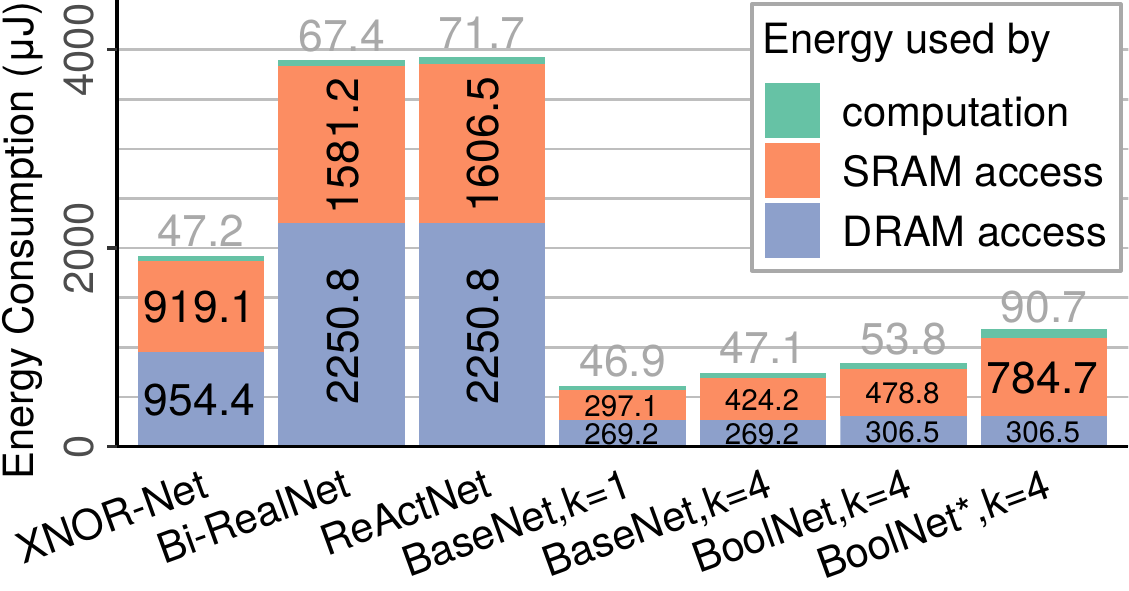}
   \caption{Energy consumption regarding computations and access to DRAM/SRAM.}
   \label{fig:sota-energy-graph}
\end{subfigure}
\end{center}
\caption{Comparison between BoolNet and state-of-the-art BNNs.
The energy consumption is calculated through hardware simulations.
BoolNet$^*$ uses dilation instead of stride in the last stage.}
\label{fig:sota}
\end{figure}
Removing 32-bit elements from previous BNNs (e.g. ReActNet \cite{liu2020reactnet}) leads to an energy reduction by up to 6$\times$ (BaseNet with $k$=1), but incurs an accuracy drop of 17\% (see Table \ref{tab:sota-table}).
Using the proposed Multi-slice strategy ($k$=4) reduces the accuracy drop by 6.2\% and still achieves 5.3$\times$ energy reduction.
Our BoolNet design further increases the accuracy by 2.5\%, but requires 12\% more energy (for a 4.5$\times$ reduction).
Compared to the result of Bi-RealNet \cite{liu2018bi}, which has been the basis for other works \cite{real2binICLR20} BoolNet with $k$=4 provides an accuracy improvement of 1.2\% (and a 4.5$\times$ energy reduction).
The accuracy of our BoolNet can be further increased with common techniques, such as replacing stride with dilation (denoted with a star$^*$) during the last stage of the network, which increases accuracy by 2\% (and yields a 3.3$\times$ reduction of energy).
Overall our results show that our proposed BaseNet and BoolNet can achieve significant energy reduction with little accuracy loss compared to recent state-of-the-art models.

\section{Conclusion}
\label{sec:conclusion}
In this paper, we studied how to balance energy consumption and accuracy of binary neural networks.
We proposed several simple yet useful strategies to remove or replace 32-bit components from BNNs.
Our novel BoolNet with fully binary information flow is constructed and still maintains reasonable accuracy.
Experiments on ImageNet and the hardware simulations show that (1) theoretical number of operations does not fully reveal the actual efficiency and (2) BoolNet is more energy-efficient with less computing requirements, lower memory usage and lower energy consumption.
We believe this is orthogonal to the goals of previous works and a meaningful first step towards achieving extremely efficient BNNs.


{\small
\bibliographystyle{ieee_fullname}
}








\appendix

\section{Appendix}

Before we present further details in the following sections, we present an overview on the total amount of computation that was used during this work.
We measured the total GPU hours for the four experiments in Section \ref{sec:sota_comparison} of our paper.
In total, all four experiments (BaseNet k=1, BaseNet k=4, BoolNet k=4, BoolNet* k=4) were trained on 4 GPUs and thus required 276, 252, 204, and 156 GPU  hours respectively, in total: 888 GPU hours.

For our ablation studies and our intermediate, initial, or discarded experiments, which were not presented in the paper, we can only provide an estimation of the amount of GPU hours, since we did not have exact measurements in place at the start of this work.
We have recorded more than 4300 GPU hours for these experimental results, but estimate that a further 1500-2000 hours were needed in the initial experiments, before we started measuring the runtime.

\subsection{Training Details and Further Experimental Results}

The training strategy is mostly based on \cite{bethge2020meliusnet}.
More specifically, we use the RAdam optimizer \cite{Liu2019radam} with a learning rate of $0.002$ without weight decay, use the \emph{cosine learning rate decay} \cite{gluoncvnlp2019}, and train with a batch size of 256 for 60 epochs.
We only use random flipping and cropping of images to a resolution of $224\times224$ for augmentation.
During validation we resize the images to $256\times256$, and then crop the center with a size of $224\times224$.
Our implementation is based on PyTorch \cite{paszke2019pytorch}, and the code can be online\footnote{\url{https://github.com/hpi-xnor/BoolNet}}.
The implementations of many previous works can not be sped up with \texttt{XNOR} and \texttt{popcount} (also observed by \cite{fromm2020riptide}), since they use padding with zeros, which introduces a third value ($\{-1,0,+1\}$) in the feature map.
To circumvent this issue, we use \emph{Replication} padding, which duplicates the outer-most values of the feature map, thus the values are limited to $\{-1,+1\}$.
A further difference to previous work, is our progressive weight binarization technique to remove the need for two-stage trainings, as discussed in the following Section.

\subsubsection{Progressive Weight Binarization vs. Two-Stage Training}

\begin{figure}[b]
\captionsetup[subfigure]{justification=centering}
\begin{center}
\includegraphics[width=\textwidth]{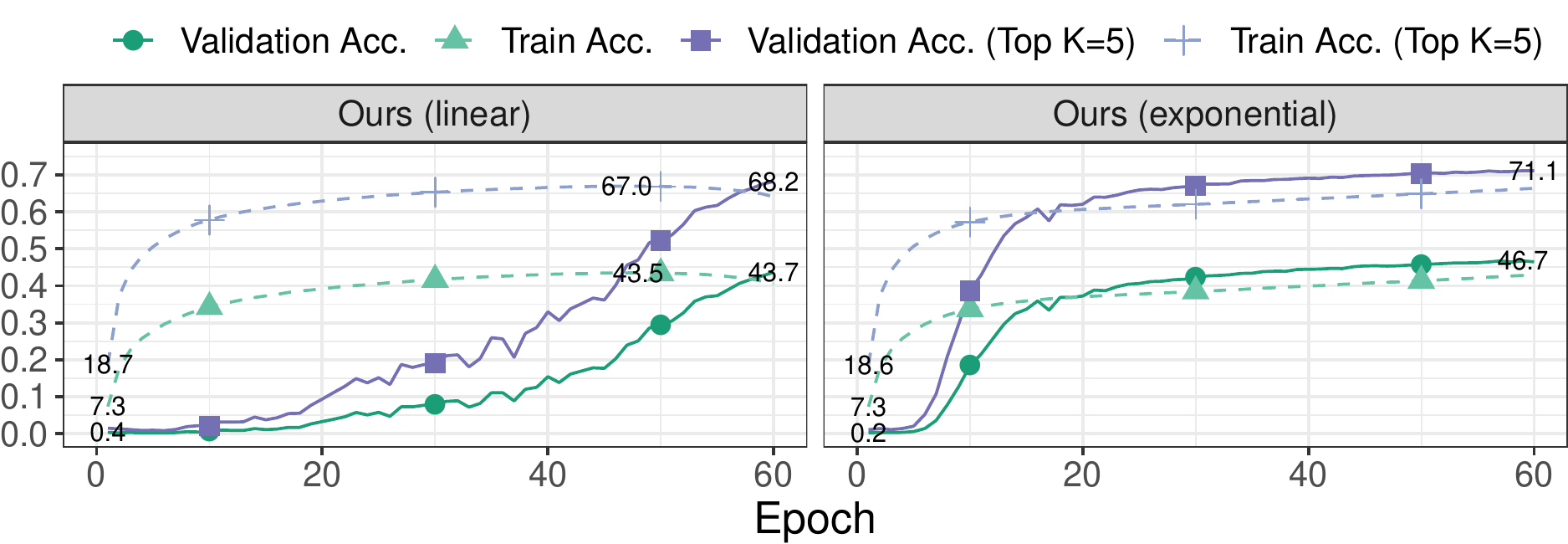}
\end{center}
\caption{The training and validation accuracy curves of our proposed \emph{Progressive Weight Binarization}.
An exponential increase of the slope leads to much better results, than a linear increase.}
\label{fig:progressive_binarization}
\end{figure}

We have introduced the progressive weight binarization strategy in Section \ref{sec:progressive-weight-binarization}, Equation \ref{equa:progressive binarizing} and discussed the results briefly in Section \ref{sec:training_details}.
As presented in our main paper, training with progressive weight binarization leads to a higher accuracy, if we train for the same total number of epochs.
However, we also conducted an experiment using a linear increase (${\lambda'}_t = 1-t+\epsilon$, $\epsilon=10^{-6}$) instead of our proposed exponential increase (${\lambda}_t = \sigma^t$) of the slope (see Figure \ref{fig:progressive_binarization}).
We chose $\sigma$, so the final $\lambda$ values are equal, i.e. if $t_\text{max}$ represents the final epoch, then $\lambda_{t_\text{max}}=\lambda'_{t_\text{max}}=10^{-6}$.
The learning curves show that our progressive weight binarization gains the largest advantage by only ``initializing'' the values during a brief initial phase of the training.


\subsubsection{Code for Reproducibility}
\label{sec:code}

We uploaded our training code and all details needed to reproduce each of our
experiments depicted in Section \ref{sec:sota_comparison} to \url{https://github.com/hpi-xnor/BoolNet}.

\subsection{Ablation Study on the Downsample Structure}

\begin{table}[]
\begin{center}

\setlength\tabcolsep{5pt}
\begin{tabular}{ccc|cc|cc|cc}
\hline
    &                    &        & \multicolumn{2}{c|}{AvgPool Acc. (\%)}   & \multicolumn{2}{c|}{MaxPool Acc. (\%)} & \multicolumn{2}{c}{Stride=2 Acc. (\%)}   \\
$k$ & {Bits} & Groups & Top-1 & Top-5 & Top-1 & Top-5 & Top-1 & Top-5 \\ \hline
\mrc{2}{1}     & 32          & 1      & 63.5 & 87.8 & 63.0 & 87.7 & 60.7 & 86.4 \\ 
               & \textbf{1}  & 1      & 63.1 & 88.0 & 62.5 & 87.2 & 60.9 & 86.7 \\ \hline
\mrc{3}{8}     & 32          & 1      & 66.0 & 89.4 & 67.0 & 90.0 & 63.4 & 87.9 \\ 
               & 32          & 8      & 65.0 & 88.0 & 65.3 & 88.9 & 62.2 & 87.0 \\
               & \textbf{1}  & 1      & 64.1 & 88.5 & 65.0 & 89.0 & 62.6 & 87.3 \\
\hline
\end{tabular}

\end{center}
\caption{Our ablation study on CIFAR100 regarding different downsampling methods.
The number of bits refers to both the input activation and weight binarization of the $1\times1$ convolution in the shortcut branch.}
\label{tab:ablation_shortcut}
\end{table}

As described in Section \ref{subsec:reducting fp ops}, we modify the $1\times1$ convolution in the downsampling branch in contrast to many previous works \cite{rastegari2016xnor,liu2018bi,liu2020reactnet,real2binICLR20}.
While being helpful for accuracy, the 32-bit $1\times1$ convolution involves extra computing, memory and energy consumption, which is in conflict with our motivation.
Using our multi-slice strategy with $k=8$, the number of input channels for the $1\times1$ convolution also increases by the same factor of 8.
To counter this increase of 32-bit operations, it could be an option to use 8 groups in the convolution, which would keep the number of 32-bit operations constant, compared to previous work.
However, this strategy still conflicts with our motivation to remove 32-bit most operations.
Furthermore, the average pooling layer used in previous work, requires additional 32-bit addition and division operations, which could be reduced with either using a max pooling layer or a stride of 2.

Therefore, to find a good downsample module with binary data flow, we first design the downsample template as [Conv$_y$, $x$, BN, Sign].
In this template, $x$ indicates the different candidate downsample operations (e.g., average pooling, max pooling, or adding stride=2 to the convolution) and $y$ the number of bits used for weights and activations in the convolution.

We conducted a detailed ablation study on the CIFAR100 dataset for both $k=1$ and $k=8$ (see Table \ref{tab:ablation_shortcut}).
The results show, that max pooling combined with 1-bit $1\times1$ convolution (groups = 1) has the same Top-1 accuracy as average pooling combined with 32-bit $1\times1$ convolution (groups = 8).
Thus, we decide to use max pooling instead of average pooling, since it does not involve any 32-bit operations, such as addition and division.

Based on the above analysis, we suggest using the [32-bit Conv (groups = k), AvgPool2d, BN, Sign] structure for the downsample branch if we want to increase accuracy. 
However, if we intend to build a fully binary data flow, we suggest using the [1-bit Conv (groups = 1), MaxPool2d, BN, Sign] structure (independent of $k$) instead to balance the accuracy and hardware efficiency.
The latter is also the structure we used for our experiments in the main paper.

\subsection{More Details About the Energy Consumption Simulation}
In Table \ref{tab:memory_all}, we give an example of calculating the memory consumption among different stages of our network.
Compared with regular BNNs with mixed precision data flow, the fully binary representation of BoolNet significantly lowers the memory consumption during inference process.
This change leads to less memory access operations to DRAM has much higher power consumption, than the on-chip SRAM.
To the best of our knowledge, our work is the first one to study the impact of memory access on energy consumption. The details of simulation and energy estimation are introduced as follow.

\begin{table}[]
\caption{Theoretical minimum memory requirement of all convolution blocks (can differ depending on the implementation).
$k$ is the number of slices.
The stages have different input size and thus lead to different memory requirements.
BoolNet$^*$ uses dilation instead of stride before the last stage, and thus needs more memory to store the features.
However BoolNet$^*$ still requires less memory than a regular BNN in the fourth stage.}
\setlength\tabcolsep{2.8pt}
\small
\begin{tabular}{c|ccc|ccc}
\Xhline{4\arrayrulewidth}
\multicolumn{1}{l|}{\multirow{2}{*}{\begin{tabular}[c]{@{}l@{}}Memory\\ Usage of\end{tabular}}} &
  \multicolumn{3}{c|}{Stage 1 with $64\times56\times56$} &
  \multicolumn{3}{c}{Stage 2 with $128\times28\times28$} \\
\multicolumn{1}{l|}{} &
  BoolNet (k=1) &
  BoolNet (k=4) &
  Regular BNN &
  BoolNet (k=1) &
  BoolNet (k=4) &
  Regular BNN \\ \hline
Weights &
  $36{,}864$ &
  $36{,}864$ &
  $36{,}864$ &
  $147{,}456$ &
  $147{,}456$ &
  $147{,}456$ \\ \hline
Activation &
  \makecell{$200{,}704{\cdot}1$\\ $= 200{,}704$} &
  \makecell{$200{,}704{\cdot}4$\\ $= 802{,}816$} &
  \makecell{$200{,}704{\cdot}1$\\ $= 200{,}704$} &
  \makecell{$100{,}352{\cdot}1$\\ $= 100{,}352$} &
  \makecell{$100{,}352{\cdot}4$\\ $= 401{,}408$} &
  \makecell{$100{,}352{\cdot}1$\\ $= 100{,}352$} \\ \hline
\makecell{Output \&\\ Features} &
  \makecell{$2{\cdot}200{,}704{\cdot}1$\\ $= 401{,}408$} &
  \makecell{$2{\cdot}200{,}704{\cdot}4$\\ $= 1{,}605{,}632$} &
  \makecell{$2{\cdot}200{,}704{\cdot}32$\\ $= 12{,}845{,}056$} &
  \makecell{$2{\cdot}100{,}352{\cdot}1$\\ $= 200{,}704$} &
  \makecell{$2{\cdot}100{,}352{\cdot}4$\\ $= 802{,}816$} &
  \makecell{$2{\cdot}100{,}352{\cdot}32$\\ $= 6{,}422{,}528$} \\ \hline
\textbf{Total} &
  \textbf{638,976} &
  \textbf{2,445,312} &
  \textbf{13,082,624} &
  \textbf{448,512} &
  \textbf{1,351,680} &
  \textbf{6,670,336} \\
  \Xhline{4\arrayrulewidth}
\end{tabular}
\small
\begin{tabular}{c|ccc|cccc}
\multicolumn{1}{l|}{\multirow{2}{*}{\begin{tabular}[c]{@{}l@{}}Memory\\Usage of\end{tabular}}} &
  \multicolumn{3}{c|}{Stage 3 with $256\times14\times14$} &
  \multicolumn{4}{c}{Stage 4 with $512\times7\times7$} \\
\multicolumn{1}{l|}{} &
  BoolNet(k=1) &
  BoolNet(k=4) &
  Regular\,BNN &
  \scriptsize BoolNet(k=1) &
  \scriptsize BoolNet(k=4) &
  \scriptsize BoolNet*(k=4) &
  \scriptsize Regular\,BNN \\ \hline
{Weights} &
  $589{,}824$ &
  $589{,}824$ &
  $589{,}824$ &
  \scriptsize $2{,}359{,}296$ &
  \scriptsize $2{,}359{,}296$ &
  \scriptsize $2{,}359{,}296$ &
  \scriptsize $2{,}359{,}296$ \\ \hline
{Activation} &
  \makecell{$50{,}176{\cdot}1$\\ $= 50{,}176$} &
  \makecell{$50{,}176{\cdot}4$\\ $= 200{,}704$} &
  \makecell{$50{,}176{\cdot}1$\\ $= 50{,}176$} &
  \makecell{\scriptsize $25{,}088{\cdot}1$\\ \scriptsize $= 25{,}088$} &
  \makecell{\scriptsize $25{,}088{\cdot}4$\\ \scriptsize $= 100{,}352$} &
  \makecell{\scriptsize $100{,}352{\cdot}4$\\ \scriptsize $= 401{,}408$} &
  \makecell{\scriptsize $25{,}088{\cdot}1$\\ \scriptsize $= 25{,}088$} \\ \hline
\makecell{Output \&\\Features} &
  \makecell{$2{\cdot}50{,}176{\cdot}1$\\ $= 100{,}352$} &
  \makecell{$2{\cdot}50{,}176{\cdot}4$\\ $= 401{,}408$} &
  \makecell{$2{\cdot}50{,}176{\cdot}32$\\ $= 3{,}211{,}264$} &
  \makecell{\scriptsize $2{\cdot}25{,}088{\cdot}1$\\ \scriptsize $= 50{,}176$} &
  \makecell{\scriptsize $2{\cdot}25{,}088{\cdot}4$\\ \scriptsize $= 200{,}704$} &
  \makecell{\scriptsize $2{\cdot}100{,}352{\cdot}4$\\ \scriptsize $= 802{,}816$} &
  \makecell{\scriptsize $2{\cdot}25{,}088{\cdot}32$\\ \scriptsize $= 1{,}605{,}632$} \\ \hline
\textbf{Total} &
  \textbf{740,352} &
  \textbf{1,191,936} &
  \textbf{3,851,264} &
  \textbf{\scriptsize 2,434,560} &
  \textbf{\scriptsize 2,660,352} &
  \textbf{\scriptsize 3,563,520} &
  \textbf{\scriptsize 3,990,016} \\
\Xhline{4\arrayrulewidth}
\end{tabular}
\label{tab:memory_all}
\end{table}

\begin{figure}[]
\captionsetup[subfigure]{justification=centering}
\begin{center}
\begin{subfigure}[t]{0.45\linewidth}
   \centering
   \includegraphics[scale=0.2]{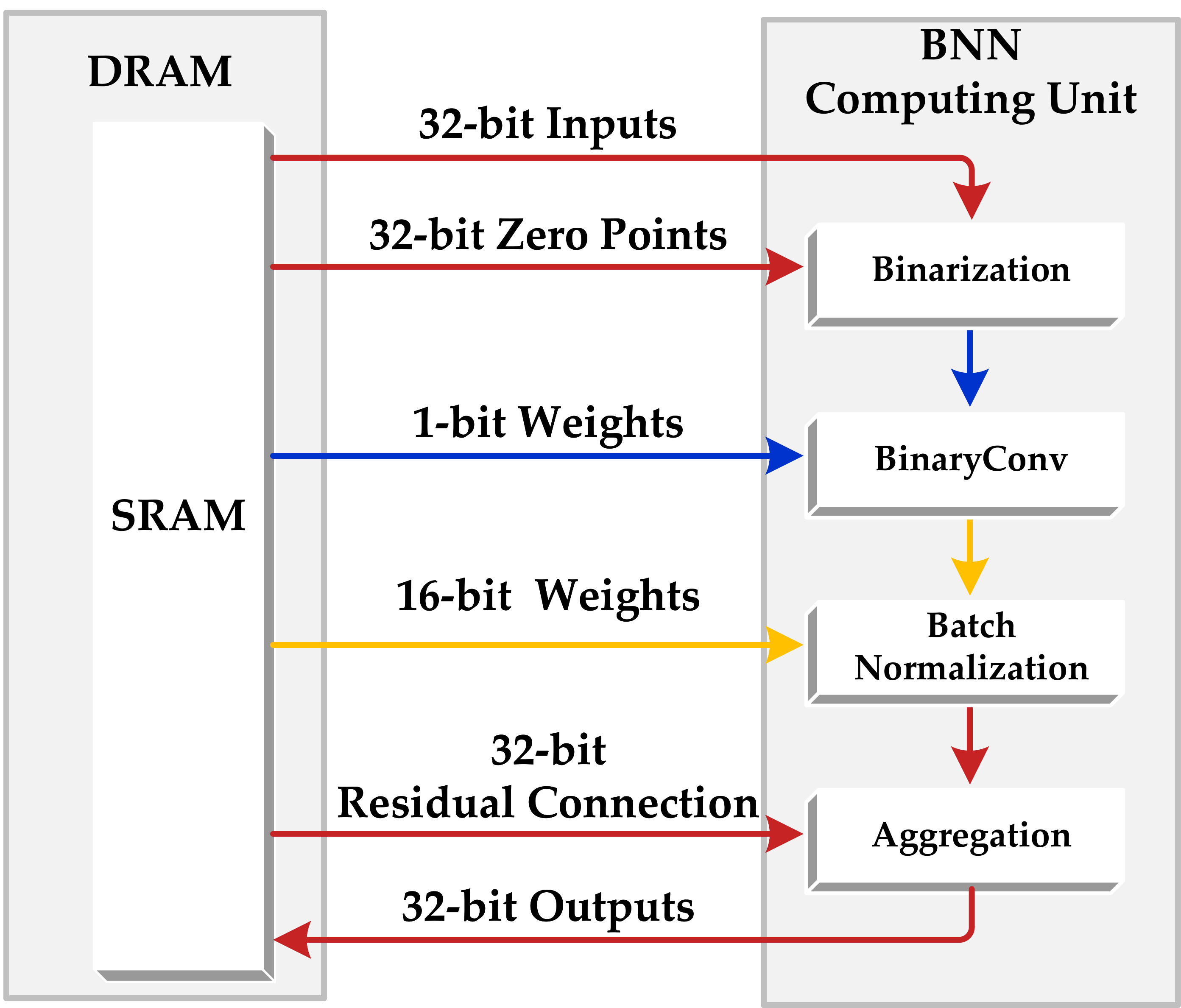}
   \caption{BiReal Net Data Flow on Hardware}
   \label{fig:hardware-birealnet}
\end{subfigure}
\hfill
\begin{subfigure}[t]{0.5\linewidth}
   \centering
   \includegraphics[scale=0.2]{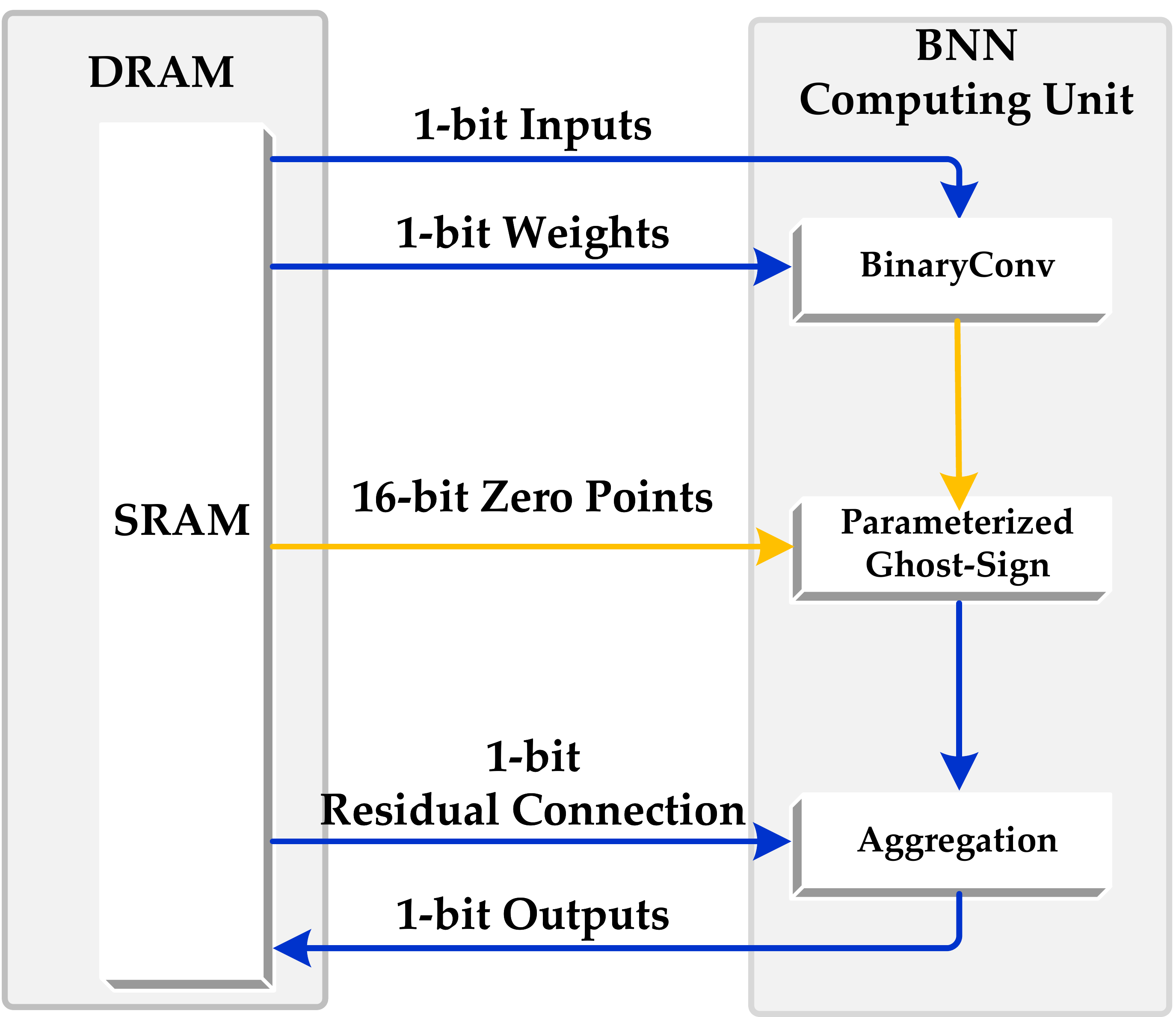}
   \caption{BoolNet Data Flow on Hardware}
   \label{fig:hardware-boolnet}
\end{subfigure}
\hfill
\end{center}
\caption{Hardware data flow comparison between BiReal Net and BoolNet.}
\label{fig:hardware data flow}
\end{figure}

\textbf{Overall architecture}. An illustrative graph on the data flow between the hardware components is provided in Figure \ref{fig:hardware data flow}. In the typical BNN Bi-RealNet, only the convolution is binary, the shortcut branch adopts high precision, and other calculations adopt high precision, too. The corresponding accelerators we designed have different computing modules (but their parallelisms are the same, that is, the computing time of the whole block is roughly the same, and the binary convolution units are exactly the same). In addition, for fair comparison, these accelerators have the same size of on-chip memory (192KB for feature map and 288KB for weight) and the same off-chip memory.

\textbf{Computing unit}. The binary convolution units of different BNN accelerators are exactly the same, but other calculation units of BoolNet are simpler. The first is the shortcut branch of downsample blocks. The shortcut branch of traditional BNNs are high-precision, and the high-precision convolution downsampling is adopted. Although the convolution on the shortcut branch accounts only for a small amount of calculation, the power consumption of a high-precision convolution is 37 times that of a binary convolution, and the extra convolution unit also increases the complexity of the circuit. Secondly, regarding batch normalization and binarization, since the shortcut branch has changed from high-precision to binary, the aggregation position of the shortcut branch and the main branch has also changed, so that the binarization and batch normalization can be simplified together, while the calculation of typical BNN can not be simplified, and their power consumption is high. In addition, there is a difference in the complexity of the aggregation operation itself (boolean logic operation vs. 32-bit addition) and the computational overhead of non-linear functions (i.e. RPReLU) added in networks such as ReActNet. These aspects show the efficiency of BoolNet. We write RTL code to realize the above design, use Design Compiler software to synthesize with a TSMC 65nm process, and simulate at 1GHz clock frequency. The software can provide the hierarchical circuit area and power of computing units, including static power (Ps) and dynamic power (Pd). For each layer of the network, we know the calculation amount (A) of each operation. According to the circuit parallelism (Pa), we can calculate the required number of cycles (Cn = A / Pa), and then calculate the energy consumption according to the frequency and power (Ec = Cn $\times$ (Ps + Pd) / 10$^-9$). 
For the operations with less calculation cycles, the energy consumption waiting for other units is estimated by static power (Es = (Cnmax - Cn) $\times$ Ps / 10$^-9$).

\textbf{On-chip memory}. We use CACTI 6.5 to simulate the power of on-chip SRAM. According to the requirements of the computing unit, we configure the on-chip SRAM to meet the parallelism of the corresponding data reading bandwidth (64 bits for BoolNet and 2048 bits for traditional BNNs), while keeping the total storage unchanged. In addition, we split a large SRAM into multiple SRAMs to meet the requirement that the read time is less than the clock cycle (1ns) of the computing unit. Finally, the simulation software can give the energy consumption of one read or one write of each SRAM unit. For each layer of the network, we know the total number of operations for each type of operation. According to the circuit parallelism, we can calculate the number of cycles. Then, according to the amount of data that needs to be read from (or written to) SRAM in each cycle, we can get the energy that the accelerator spends to access on-chip SRAM.

\textbf{Off-chip memory}. Due to the limited amount of on-chip memory, it is inevitable to save some data to (or read from) off-chip DRAM in BNN computing. In our BoolNet design, due to the large total number of weights, all BNN accelerators need to read weights from DRAM and write to SRAM before the computation of each layer. In addition, for traditional BNN, the intermediate feature maps are larger, which cannot be completely cached on-chip. It is also necessary to save the extra part to DRAM, to read it back in the next layer. With the amount of read-write operations of data to (and from) DRAM and SRAM, the power consumption data of DRAM read-write operations (SRAM has been given by the CACTI simulation in the previous step) is also needed to estimate the overall energy consumption. We use the DDR4 Power Calculator provided by Micron, to configure a DDR UDIMM module composed of four 8Gb x16 chips, which adopts the speed grade of -075E, and the maximum transmission rate is 2666MT/s. The calculator gives the average energy consumption of reading and writing data with 64 bits parallelism.





\end{document}